\documentclass[letterpaper, 10 pt, conference]{ieeeconf}  

\IEEEoverridecommandlockouts                              

\usepackage{booktabs}
\usepackage{graphicx}
\usepackage{amsmath}
\usepackage{amssymb}
\usepackage{cite}
\usepackage{stfloats}
\title{\LARGE \bf
CAR-LOAM: Color-Assisted Robust LiDAR Odometry and Mapping}
\author{Yufei Lu, Yuetao Li, Zhizhou Jia, Qun Hao, and Shaohui Zhang
\thanks{All authors are with the School of Optics and Photonics, Beijing Institute of Technology, Beijing 100081, China. (Corresponding author: Shaohui Zhang.)
        {\tt\small \{luyufei, liyuetaochn, jiazhizhou, qhao, zhangshaohui\}@bit.edu.cn}}%
}

\begin{document}

\maketitle 
\thispagestyle{empty}
\pagestyle{empty}
\begin{abstract}
In this letter, we propose a color-assisted robust framework for accurate LiDAR odometry and mapping (LOAM). Simultaneously receiving data from both the LiDAR and the camera, the framework utilizes the color information from the camera images to colorize the LiDAR point clouds and then performs iterative pose optimization. For each LiDAR scan, the edge and planar features are extracted and colored using the corresponding image and then matched to a global map. Specifically, we adopt a perceptually uniform color difference weighting strategy to exclude color correspondence outliers and a robust error metric based on the Welsch’s function to mitigate the impact of positional correspondence outliers during the pose optimization process. As a result, the system achieves accurate localization and reconstructs dense, accurate, colored and three-dimensional (3D) maps of the environment. Thorough experiments with challenging scenarios, including complex forests and a campus, show that our method provides higher robustness and accuracy compared with current state-of-the-art methods.
\end{abstract}
\section{INTRODUCTION}
Light Detection and Ranging (LiDAR) has become one of the most critical perception modalities in robotic systems owing to its high accuracy, long range, and reliability. By enabling state estimation in six degrees of freedom (DoF) and construction of precise maps of the surrounding environment, LiDAR-based Simultaneous Localization and Mapping (SLAM) has found applications in autonomous driving \cite{5940562}, drone inspection \cite{8461191}, logistics \cite{ito2018small}, and other areas. Iterative Closest Point (ICP) \cite{121791} is the most classical method to achieve state estimation by iteratively minimizing the distance between two consecutive LiDAR scans to derive the transformation. However, it has a heavy computation load since a large number of points are involved in optimization.
Lidar Odometry And Mapping (LOAM) \cite{zhang2017low} alleviates the computation load by matching only the extracted edge and planar features. As LiDAR technology continues to advance, solid-state LiDARs are increasingly preferred for their low cost, lightweight design and high resolution—advantages that traditional multi-line spinning LiDARs do not possess. Following this trend, methods based on solid-state LiDARs such as Loam-Livox \cite{9197440} and SSL-SLAM \cite{9357899} have been developed, which extend the applicability of LOAM. However, as a key part of LOAM, scan matching is highly susceptible to correspondence outliers \cite{WOS:000323204000012} (both color and positional). Specifically, during each scan matching process, a large number of closest-point pairs must first be established. These closest-point pairs serve as inputs to the pose optimization submodule, and their quality directly affects the accuracy of pose estimation. A substantial number of correspondence outliers can cause the pose optimization to proceed in a significantly biased direction. Consequently, as poses are estimated sequentially over time, odometry drift accumulates, leading to a significant reduction in accuracy. In addition, existing LOAM algorithms generally lack mechanisms for mitigating the impact of outliers. Therefore, we propose an accurate LOAM framework CAR-LOAM that introduces a perceptually uniform color difference weighting strategy to exclude color correspondence outliers and a robust error metric based on the Welsch’s function \cite{holland1977robust} to suppress positional correspondence outliers. The main contributions of our work are:
\begin{figure}[t]
    \centering
    \includegraphics[width=1\linewidth]{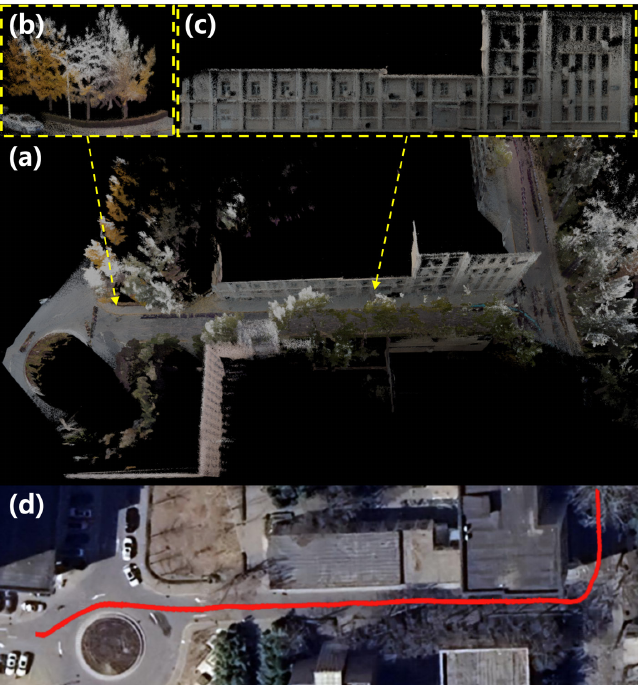}
    \caption{
    (a) A dense, accurate, colored and 3D point cloud map reconstructed by CAR-LOAM.
    (b) A close-up of (a), highlighting autumn trees and a car beneath them.
    (c) Another close-up of (a), showing one side of the building, including its doors and windows.
    (d) A corresponding satellite image of the campus environment, with the red path indicating the traveling trajectory of our mobile platform.
    }
    \label{fig:1}
\end{figure}
\begin{itemize}
\item We develop a color-assisted robust framework for accurate LOAM that uses color from camera images exclusively to assist the LiDAR odometry rather than incorporating an additional visual pose estimation module.
\item We propose a perceptually uniform color difference weighting strategy based on the CIEDE2000 formula \cite{WOS:000225857800003} and a Gaussian robust function, which weakly penalizes point pairs with large color differences during scan matching.
\item We introduce a robust error metric based on the Welsch's function in the scan matching process, which can be considered a soft thresholding approach for suppressing positional correspondence outliers.
\item We evaluate our method on datasets and in real-world experiments, demonstrating its ability to accurately localize and generate dense, accurate, colored and 3D point cloud maps (see Fig. 1 and Fig. 12).
\end{itemize}
\section{RELATED WORK}
Here, we review existing research closely related to our work. First, we summarize LiDAR-only odometry and mapping, categorized into direct and feature-based methods. We then provide an overview of loosely-coupled and tightly-coupled LiDAR-Inertial odometry methods.
\subsection{LiDAR-only Odometry and Mapping}
\emph{1) Direct Methods:}
The most prominent methods of scan matching in LiDAR-based SLAM rely on the ICP \cite{121791} paradigm. Starting from an initial alignment, ICP alternates between finding the closest point pairs between two point clouds and iteratively minimizing the distances of correspondences to get the optimal transformation parameters. With ICP, multiple LiDAR scans are aligned to generate a local map and output the corresponding poses, forming a direct LOAM module. During this process, the method operates on raw data, with or without using spatial downsample or temporal downsample. To reduce the mismatch rate, in addition to using point-to-point distance in traditional ICP, various error metrics, such as point-to-plane \cite{132043}, can be applied in direct LOAM methods. While accounting for all environmental details, direct methods are computationally expensive and also introduce more outliers and noises.

\emph{2) Feature-based Methods:}
To improve accuracy and reduce computation load, many feature-based methods match only edge and planar features. The introduction of LOAM \cite{zhang2017low} establishes a standard for numerous subsequent works. For each incoming scan, edge and planar features are pre-extracted through local smoothness analysis and then aligned with the preceding scan to obtain odometry. With the odometry, multiple scans form a sweep, which is then registered to a global map incrementally. The cost in optimization is calculated as the Euclidean distance between points and their corresponding edges and surfaces.
To enable the algorithm to run in real-time on computationally constrained platforms, LeGO-LOAM \cite{8594299} separates ground plane optimization before feature extraction. To further enhance running performance and accuracy, F-LOAM \cite{9636655} adopts a non-iterative two-stage distortion compensation method and solves a weighted nonlinear least squares problem to achieve scan-to-map matching. During the pose optimization, a weight function based on local smoothness is introduced to increase the penalty on points with lower curvature in edge features and points with higher curvature in planar features to balance the matching process. 
The above works \cite{8594299, 9636655, zhang2017low} mainly focus on multi-line spinning LiDARs. To solve the problem of algorithm adaptation for solid-state LiDARs, LOAM-Livox \cite{9197440} considers the scanning mechanisms and low-level physical properties of solid-state LiDARs and performs point-level selection to extract reliable features. However, existing algorithms generally lack mechanisms for outlier suppression, which limits their accuracy. 
\subsection{LiDAR-Inertial Odometry}
Since pure LiDAR odometry often incurs drift over long-term operation and degenerates in featureless environments, fusion with the Inertial Measurement Unit (IMU) is an elegant solution. Current works on LiDAR-Inertial fusion can be categorized into two classes: loosely coupled and tightly coupled.

\emph{1) Loosely-coupled Methods:}
Loosely-coupled methods commonly handle LiDAR and IMU data independently and fuse their results in a modular fashion. In addition to LiDAR-only methods, \cite{zhang2017low, 8594299} also provide typical examples of loosely-coupled IMU-aided LiDAR odometry, where poses integrated from IMU data serve as the initial guesses for LiDAR scan registration.

\emph{2) Tightly-coupled Methods:}
Tightly-coupled LiDAR-inertial odometry can be divided into optimization-based and filter-based approaches. In various optimization processes (e.g., sliding window joint optimization \cite{8793511, 9392274}, and factor graph smoothing \cite{9341176}), optimization-based approaches account for the intrinsic characteristics of the two sensors, such as IMU observation noise, biases, and residuals from LiDAR scan registration, rather than simply fusing the final results.
For filter-based approaches, LINS \cite{9197567} is the first tightly-coupled method that achieves the 6 DOF ego-motion estimation using the iterated Kalman filter. To further enhance computational efficiency and lower the odometry drift, FAST-LIO \cite{9372856} proposes a new formula to compute the Kalman gain and a back-propagation to resolve motion distortion. In its subsequent work FAST-LIO2 \cite{9697912}, an incremental iKD-Tree map is maintained, significantly accelerating the mapping process.
Furthermore, with regard to map representation, ImMesh \cite{10304337}, building on the work of \cite{9813516}, becomes the first tightly-coupled LiDAR-inertial framework capable of reconstructing triangle meshes of large-scale environments online without relying on GPU acceleration.
Although fusion with IMU can achieve higher accuracy in long-range and degenerate scenarios, it introduces complex calibration and substantial computation.

Our method falls into the feature-based LOAM. We adopt scan-to-map matching similar to \cite{9197440}, but with a significant suppression of color and positional correspondence outliers, requiring only the addition of color image input. 
\begin{figure*}[!ht] 
    \vspace{0.3cm}
    \centering
    \includegraphics[width=7in]{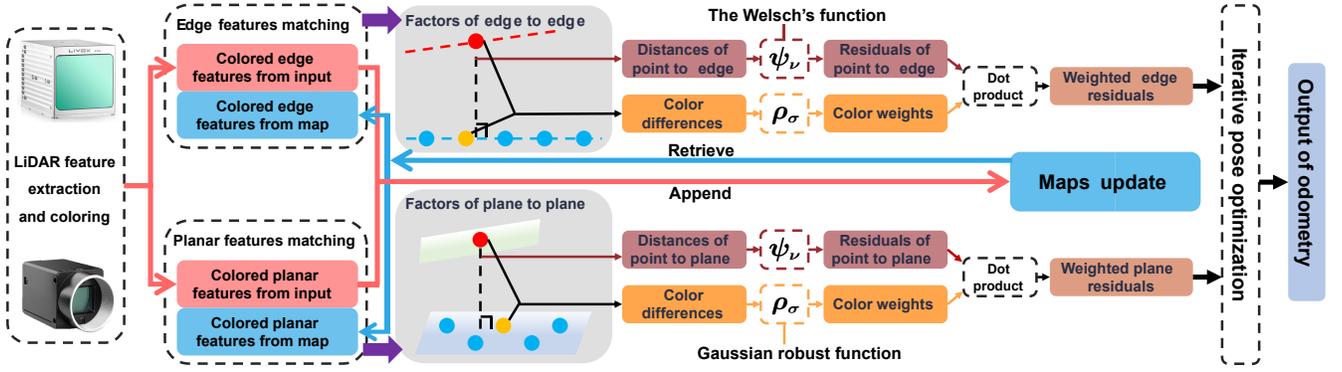} 
    \caption{The overview of our workflows. Each new LiDAR frame, colored by the corresponding RGB color image from the camera, is matched with the global map to provide the odometry output. The matching result is in turn used to register the frame to the global map. During the matching process, each point on the new colored edge/planar feature is used to calculate the distance and perceptually uniform color difference to its corresponding edge/planar feature retrieved from the global map. The color of the edge/planar feature is represented by the color of the nearest neighbor point of the current query point. The distances serve as inputs to the Welsch's function to get robust normalized residuals. Similarly, perceptually uniform color differences are processed using the Gaussian robust function to generate normalized color weights. The residuals are then multiplied by their corresponding weights and utilized in iterative pose optimization.}
    \label{fig:2}
\end{figure*}
\begin{figure}[t]
    \centering
    \includegraphics[width=3.4in]{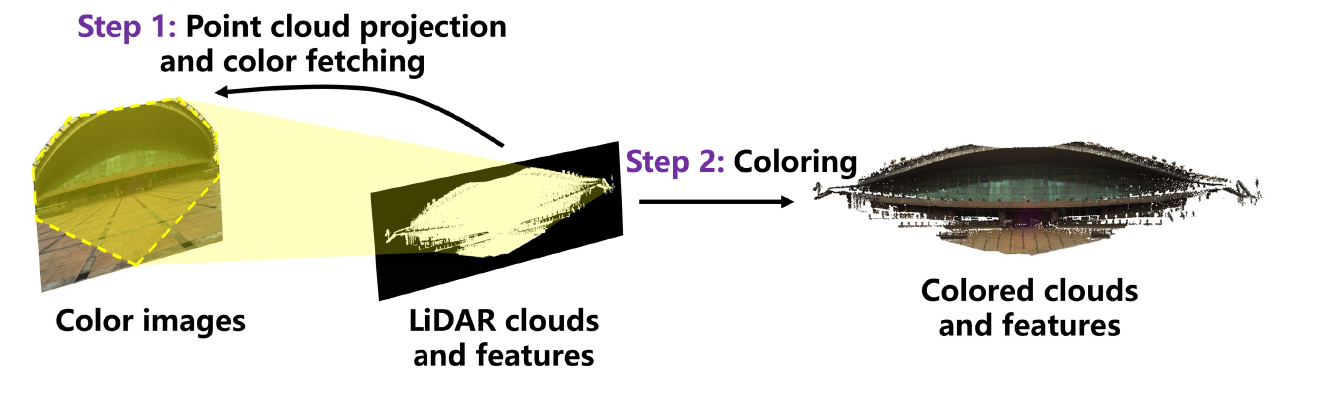}
    \caption{Steps of the coloring of LiDAR clouds and features.}
    \label{fig:3}
\end{figure}
\section{LIDAR FEATURE EXTRACTION AND COLORING}
As a LiDAR-based odometry and mapping system, our method uses images solely to provide color information about the scene and does not introduce any additional visual odometry module. The LiDAR and camera have been pre-calibrated. During system operation, each LiDAR scan is paired with the color image that has the closest timestamp to it. The overview of our workflows is shown in Fig.~\ref{fig:2}, whose front-end processing consists of the LiDAR feature extraction and coloring. 

To ensure robustness and efficiency in practice, our method selects reliable points as in \cite{9197440} and extracts edge and planar features from LiDAR scans based on local smoothness analysis as in \cite{zhang2017low}. 

After feature extraction, the raw LiDAR clouds and extracted features are colored as shown in Fig.~\ref{fig:3}. The raw LiDAR clouds are not involved in the computations and are only used for map visualization. In Step 1, each 3D point in the LiDAR clouds and features is first projected onto the raw color image plane using the pre-calibrated extrinsic and intrinsic parameters to obtain its corresponding two-dimensional (2D) pixel coordinates. Then, the RGB color at each pixel is fetched. In Step 2, we color the LiDAR clouds and features using the RGB-color information obtained in Step 1. During this process, only the 3D points with projected 2D coordinates falling within the image boundaries are retained. 

To project LiDAR points onto the image plane, they are initially transformed from the LiDAR coordinate system $\{L\}$ to the camera coordinate system $\{C\}$. Taking a point $^{L}{\mathbf{p}} = (^{L}x,^{L}y,^{L}z)$ as an example, with the intensity component omitted. This transformation is accomplished by left-multiplying the point with the 4×4 extrinsic matrix from the LiDAR to the camera, denoted $T_{CL}$.
\begin{equation}
^{C}\mathbf{p} = T_{CL}{^{L}\mathbf{p}}\tag{1}
\label{eq:transformation1}
\end{equation}
Then, using the z-axis component of $^{C}\mathbf{p}$, we project the point onto the normalized plane.
\begin{equation}
^{C}\mathbf{p} = \frac{1}{^{C}z}{^{C}\mathbf{p}}\tag{2}
\label{eq:transformation2}
\end{equation}
\begin{figure}[t]
    \centering
    \includegraphics[width=3.4in]{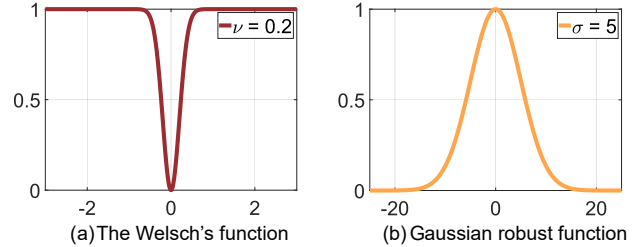}
    \caption{The Welsch's function $\psi_\nu$ with $\nu=2$ and the Gaussian robust function $\rho_\sigma$ with $\sigma=5$. }
    \label{fig:enter-label}
\end{figure}
Due to distortions inherent in the camera, the coordinates of $^{C}\mathbf{p}$ after distortion are calculated based on \cite{888718} and denoted $^{C}\mathbf{p}_{distorted}$. The corresponding pixel coordinates $\mathbf{p}_{uv}$ are computed using (3).
\begin{equation}
\mathbf{p}_{uv}=
\begin{pmatrix}
    u \\
    v \\
    1
\end{pmatrix} = K\cdot{^{C}\mathbf{p}_{distorted}}\tag{3}
\label{eq:transformation3}
\end{equation}
where the camera intrinsic matrix is denoted $K$. Next, we extend $^{L}\mathbf{p}$ with the RGB color at the image coordinates $(u, v)$ to obtain $ ^{L}\mathbf{p} = \left(^{L}x,\ ^{L}y,\ ^{L}z,\ \text{intensity},\ r,\ g,\ b\right)$.
After the LiDAR clouds and features are colored, they proceed to the next Iterative Pose Optimization module.
\section{ITERATIVE POSE OPTIMIZATION}
The previous section has described the system's data preprocessing, including LiDAR feature extraction and coloring. Next, the extracted color features are matched to the global map to obtain the odometry output and incrementally expand the map.
The pose estimation aligns the current edge features $\boldsymbol{\mathcal{E}}_k$ and planar features $\boldsymbol{\mathcal{H}}_k$ in the LiDAR coordinate system with the global feature map. Denote $\boldsymbol{\mathcal{E}}_m$ and $\boldsymbol{\mathcal{H}}_m$ the set of all edge features and planar features in the world coordinate system in the map. 
Traditional methods directly add point-to-edge and point-to-plane distances to optimization. For each edge feature point ${\mathbf{p}}_{\mathcal{E}_{k}} \in \boldsymbol{\mathcal{E}}_k$, it is first transformed from the LiDAR frame to the global frame ${}^G\mathbf{p}_e=\mathbf{T}_k\mathbf{p}_{\mathcal{E}_{k}}$, where $\mathbf{T}_k$ represents the current LiDAR pose. Let $\mathbf{p}_{i{\mathcal{E}_{m}}} \in \boldsymbol{\mathcal{E}}_m$ denote the $i$-th nearest points of ${}^G\mathbf{p}_e$ in the global map. Then the covariance matrix $ \boldsymbol{\Sigma}_\mathcal{E} $ formed by the $ n $ (i.e., 5) nearest points of $ {}^G\mathbf{p}_e $ is computed. If the biggest of eigenvalue of $ \boldsymbol{\Sigma}_\mathcal{E} $ is large enough \cite{9197440}, assuring that the nearest points of $ {}^G\mathbf{p}_e $ form a line on which $ {}^G\mathbf{p}_e $ should lie. The corresponding point-to-edge distance is then given by:
\begin{align*}\mathit{d}_{p2e}\left({}^G\mathbf{p}_e \right) & = \frac{\left\| ({}^G\mathbf{p}_e - \mathbf{p}_{1\mathcal{E}_m}) \times (\mathbf{p}_{n\mathcal{E}_m} - \mathbf{p}_{1\mathcal{E}_m}) \right\|}{\left\| \mathbf{p}_{n\mathcal{E}_m} - \mathbf{p}_{1\mathcal{E}_m} \right\|}\\
&=\frac{\left\| ({}^G\mathbf{p}_e - \mathbf{p}_{1\mathcal{E}_m}) \times \mathbf{n}_e \right\|}{\left\| \mathbf{n}_e \right\|}\\ 
&= \left\| ({}^G\mathbf{p}_e - \mathbf{p}_{1\mathcal{E}_m}) \times \hat{\mathbf{n}_e} \right\|\tag{4}\end{align*}
where $\mathbf{n}_e$ serves as a direction vector of this line and $\hat{\mathbf{n}_e}$ is the unit vector of $\mathbf{n}_e$.

Similarly, for each planar feature point ${\mathbf{p}}_{\mathcal{H}_k} \in \boldsymbol{\mathcal{H}}_k$, after obtaining ${}^G\mathbf{p}_h=\mathbf{T}_k\mathbf{p}_{\mathcal{H}_k}$, the covariance matrix $ \boldsymbol{\Sigma}_{\mathcal{H}} $ formed by the $ n $ nearest points of $ {}^G\mathbf{p}_h $ is computed. Let $\mathbf{p}_{i{\mathcal{H}_{m}}}$ denote the $i$-th nearest points of ${}^G\mathbf{p}_h$ in the global map. Note that different from the edge, if the smallest of eigenvalue of $ \boldsymbol{\Sigma}_{\mathcal{H}} $ is small enough \cite{9197440}, computing the distance of $ {}^G\mathbf{p}_h $ and the plane formed by its $ n $ (i.e., 5) nearest neighbors.
\begin{align*}
&\mathit{d}_{p2h}\left({}^G\mathbf{p}_h\right) \\ 
&= 
\frac{\left({}^G\mathbf{p}_h -\mathbf{p}_{1\mathcal{H}_m}\right)^T\left(\left(\mathbf{p}_{3\mathcal{H}_m}-\mathbf{p}_{n\mathcal{H}_m}\right)\times\left(\mathbf{p}_{3\mathcal{H}_m}-\mathbf{p}_{1\mathcal{H}_m}\right)\right)}{\|\left(\mathbf{p}_{3\mathcal{H}_m}-\mathbf{p}_{n\mathcal{H}_m}\right)\times\left(\mathbf{p}_{3\mathcal{H}_m}-\mathbf{p}_{1\mathcal{H}_m}\right)\|} \\ 
&= \left({}^G\mathbf{p}_h -\mathbf{p}_{1\mathcal{H}_m}\right)^T \frac{\mathbf{n}_h}{\|\mathbf{n}_h\|} \\
&= \left({}^G\mathbf{p}_h -\mathbf{p}_{1\mathcal{H}_m}\right)^T \hat{\mathbf{n}_h}
\tag{5}\end{align*}
where $\mathbf{n}_h$ serves as a normal of this plane and $\hat{\mathbf{n}_h}$ is the unit vector of $\mathbf{n}_h$.

Traditional solutions directly optimize the geometric distances mentioned above without considering the impact of numerous correspondence outliers from cluttered environments. However, the scan matching process is essentially a nonlinear least squares problem and is highly sensitive to outliers and noises. Measuring the alignment error using the Euclidean distance, which penalizes every point pair with a large distance in the current scan to the global map, may lead to erroneous alignment in the presence of positional correspondence outliers. We resolve this issue by adopting a robust error metric based on the Welsch’s function. Specifically, we formulate the point-to-edge residual as
\begin{equation}
\mathit{r}_{p2e}\left({}^G\mathbf{p}_e \right)=\psi_\nu(\mathit{d}_{p2e}\left({}^G\mathbf{p}_e \right))\tag{6}
\end{equation}
and the point-to-plane residual as
\begin{equation}
\mathit{r}_{p2h}\left({}^G\mathbf{p}_h \right)=\psi_\nu(\mathit{d}_{p2h}\left({}^G\mathbf{p}_h \right))\tag{7}
\end{equation}
where $\psi_\nu$ is the Welsch’s function \cite{holland1977robust}:
\begin{equation}\psi_\nu(x)=1-\exp\Bigl(-\frac{x^2}{2\nu^2}\Bigr)\tag{8}\end{equation}
and $\nu >$ 0 is a user-specified parameter.
Fig. 4 (a) presents the graphs of $\psi_\nu$ with $\nu=0.2$, which is the $\nu$ value used in our work. Since \( \psi_\nu(x) \) is monotonically increasing on \( [0, +\infty) \) and upper-bounded by 1, it effectively excludes point pairs with large positional differences, while using a Gaussian weight that gradually decreases with increasing distance. This method weakly penalizes positional correspondence outliers, leading to more stable results. Consequently, our metric via the Welsch's function is not sensitive to large deviations caused by positional correspondence outliers and noises, thereby improving the accuracy of pose estimation.

Although we use a robust metric based on the Welsch's function to reduce the impact of positional correspondence outliers, it is not sufficient for achieving accurate pose estimation. During registration, while point pairs with large positional differences are suppressed, outliers with significant color differences may still drive the optimization in the wrong direction. To exclude point pairs with large color differences and further refine the accuracy of pose estimation, we propose a weighting strategy based on the perceptually uniform color difference and the Gaussian robust function. while calculating $\mathit{r}_{p2e}\left({}^G\mathbf{p}_e \right)$, we also compute the color difference between $ {}^G\mathbf{p}_e $ and its corresponding edge feature. The edge feature's color is represented by $ \mathbf{p}_{1\mathcal{E}_m} $, which is the nearest neighbor of $ {}^G\mathbf{p}_e $. The color difference is calculated in the $\mathrm{L}^*\mathrm{C}^*\mathrm{H}^*$ color space using the CIEDE2000 formula \cite{WOS:000225857800003} for perceptual uniformity, denoted as $ \Delta E_{00}^*{_{({}^G\mathbf{p}_e,\ \mathbf{p}_{1\mathcal{E}_m})}} $. 
Note that conversion of the points from RGB to $\mathrm{L}^*\mathrm{C}^*\mathrm{H}^*$ color space is performed here. The color weight is then determined as 
\begin{equation} W{\left({}^G\mathbf{p}_e\right)} = \rho_\sigma\left(\Delta E_{00}^*{_{({}^G\mathbf{p}_e,\ \mathbf{p}_{1\mathcal{E}_m})}}\right)\tag{9}\end{equation}
similarly, the color weight of point to plane is determined as
\begin{equation} W{\left({}^G\mathbf{p}_h\right)} = \rho_\sigma\left(\Delta E_{00}^*{_{({}^G\mathbf{p}_h,\ \mathbf{p}_{1\mathcal{H}_m})}}\right)\tag{10}\end{equation}
where $\rho_\sigma$ is the Gaussian robust function:
\begin{equation}
    \rho_\sigma(x) = \exp\left(-\frac{x^2}{2\sigma^2}\right)\tag{11}
\end{equation}
and $\sigma >$ 0 is a user-specified parameter. Fig. 4 (b) presents the graphs of $\rho_\sigma$ with $\sigma=5$, which is the $\sigma$ value used in our work.
The new pose is estimated by minimizing the weighted sum of the point-to-edge and point-to-plane residuals: 
\begin{equation}\mathop {\min }\limits_{{{\mathbf{T}}_k}} \sum W{\left({}^G\mathbf{p}_e\right)}{\mathit{r}_{p2e}\left({}^G\mathbf{p}_e \right)} + \sum W{\left({}^G\mathbf{p}_h\right)}{\mathit{r}_{p2h}\left({}^G\mathbf{p}_h \right)}\tag{12}\end{equation}
For a minimal representation of the transformation, We use the Lie algebra $\mathfrak{se}(3)$ corresponding to the tangent space of $SE(3)$ at the identity, which is free from the singularities of gimbal locks \cite{Diebel2006RepresentingA}. Denoting the algebra elements $ \boldsymbol\xi = (\mathbf{v}, \boldsymbol{\omega})^T \in \mathbb{R}^6 $, where $ \mathbf{v} $ is referred to as the linear velocity and $ \boldsymbol{\omega} $ the angular velocity. The elements $ \boldsymbol\xi $ are mapped to $ SE(3) $ by the exponential map \cite{ma2005invitation}.
\begin{equation}
    {\mathbf T}(\boldsymbol\xi)=\exp(\boldsymbol{{\xi}^{\wedge}})\tag{13}
\end{equation}
Since (12) is nonlinear in $\mathbf{T_k}$, we solve it through the Gauss-Newton method. The Jacobian can be estimated by applying the left perturbation model with $\delta \xi \in \mathfrak{se}(3)$ \cite{barfoot2017state}:
\begin{align*}{{\mathbf{J}}_p} & = \frac{{\partial {\mathbf{Tp}}}}{{\partial \delta \xi }} = \mathop {\lim }\limits_{\delta \xi \to {\mathbf{0}}} \frac{{(\exp (\delta \xi ){\mathbf{Tp}} - {\mathbf{Tp}})}}{{\delta \xi }} \\ & = \left[ {\begin{array}{cc} {{{\mathbf{I}}_{3 \times 3}}}&{ - {{[{\mathbf{Tp}}]}_ \times }} \\ {{{\mathbf{0}}_{1 \times 3}}}&{{{\mathbf{0}}_{1 \times 3}}} \end{array}} \right]\tag{14}\end{align*}
where $ {[{\mathbf{T}_k \mathbf{p}_k}]}_\times $ includes both the conversion of $ \mathbf{p}_k $ from homogeneous to non-homogeneous coordinates and the calculation of the skew-symmetric matrix of $ {\mathbf{T}_k \mathbf{p}_k} $. The Jacobian matrix of the weighted edge residual can be calculated by:
\begin{align*}
{{\mathbf{J}}_{\mathcal{E}}} & = W{\left({}^G\mathbf{p}_e\right)}\frac{{\partial {\mathit{r}_{p2e}}}}{\partial{\psi_\nu}}\frac{\partial{\psi_\nu}}{\partial{d_{p2e}}}\frac{\partial{d_{p2e}}}{{\partial {\mathbf{Tp}}}}\frac{{\partial {\mathbf{Tp}}}}{{\partial \delta \xi }} \\
& = W{\left({}^G\mathbf{p}_e\right)}\frac{{d_{p2e}}}{\nu^2}\exp\left(-\frac{d_{p2e}^2}{2\nu^2}\right)\mathbf{p}_n[\hat{\mathbf{n}}_e]_{\times}\, \mathbf{p}_{\mathcal{E}_k}^\top
 {\mathbf{J}}_p\tag{15}
\end{align*}
where $\mathbf{p}_{\mathcal{E}_k}$ is the point on the edge feature of the current LiDAR frame, as described at the beginning of this section. $\mathbf{p}_n$ is the unit vector given by
\begin{align*}
    \mathbf{p}_n=\left( \frac{ \left( \mathbf{T}_k\, \mathbf{p}_{\mathcal{E}_k} - \mathbf{p}_{1\mathcal{E}_m} \right) \times \hat{\mathbf{n}}_e }{ \left\| \left( \mathbf{T}_k\, \mathbf{p}_{\mathcal{E}_k} - \mathbf{p}_{1\mathcal{E}_m} \right) \times \hat{\mathbf{n}}_e \right\| } \right)^\top\tag{16}
\end{align*}
Similarly, we can derive
\begin{align*}
{{\mathbf{J}}_{\mathcal{H}}} & = W{\left({}^G\mathbf{p}_h\right)}\frac{{\partial {\mathit{r}_{p2h}}}}{\partial{\psi_\nu}}\frac{\partial{\psi_\nu}}{\partial{d_{p2h}}}\frac{\partial{d_{p2h}}}{{\partial {\mathbf{Tp}}}}\frac{{\partial {\mathbf{Tp}}}}{{\partial \delta \xi }} \\
& = W\left( {}^G\mathbf{p}_h \right) \frac{ d_{p2h} }{ \nu^2 } \exp\left( -\frac{ d_{p2h}^2 }{ 2\nu^2 } \right) \hat{\mathbf{n}}_h^\top\, \mathbf{J}_p\tag{17}
\end{align*}
The estimation of the entire odometry can be derived by solving the nonlinear optimization iteratively.
\section{EXPERIMENTS AND RESULTS}
We perform dataset and real-world experiments to evaluate the proposed CAR-LOAM system. In the first experiment, we conduct robustness tests on public datasets from complex forest environments. Subsequently, we carry out field tests in a large-scale outdoor campus environment. The proposed method is compared against four state-of-the-art algorithms using ground truth data generated by the Real-time Kinematic Positioning (RTK) system. Furthermore, we conduct a detailed numerical analysis to quantify the accuracy of our method. Additionally, we perform an application-level experiment to evaluate the system's capability to reconstruct the 3D appearance of a large building.
\begin{figure}[t]
    \vspace{0.3cm}
    \centering
    \includegraphics[width=1\linewidth]{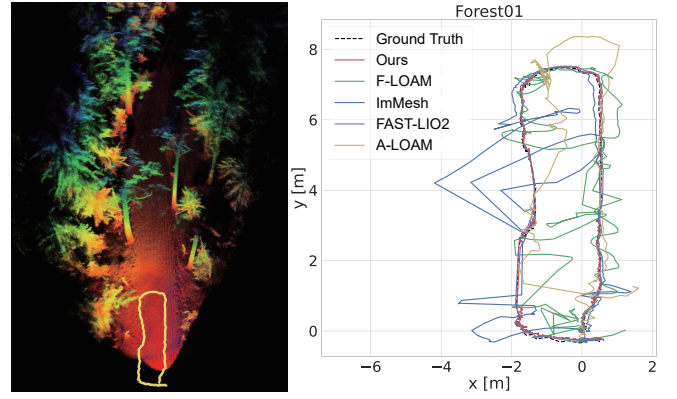}
    \caption{Left: mapping results in Forest01 by CAR-LOAM. The yellow path is the corresponding trajectory. Right: Trajectories in Forest01.}
    \label{fig:5}
\end{figure}
\begin{figure}[t]
    \centering
    \includegraphics[width=1\linewidth]{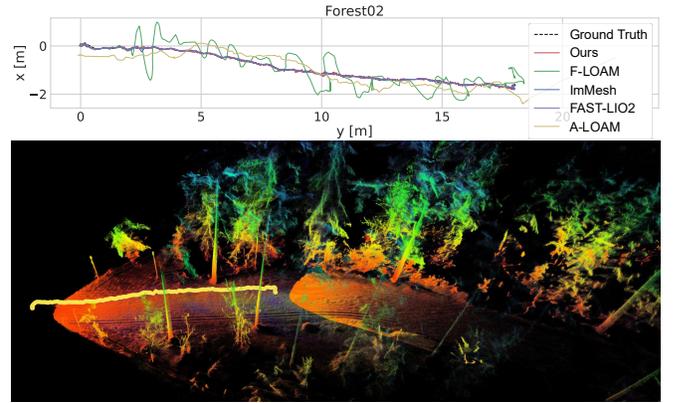}
    \caption{Left: mapping results in Forest02 by CAR-LOAM. The yellow path is the corresponding trajectory. Right: Trajectories in Forest02.}
    \label{fig:6}
\end{figure}
\subsection{Experiment-1: Robustness Evaluation in complex forest environments}
To evaluate the robustness of our method, We use the tiers-lidars-dataset-enhanced \cite{sier2023benchmark}, which include two distinct and challenging forest sequences (Forest01 and Forest02) characterized by dense vegetation, foliage occlusion, and numerous noises. Notably, both sequences are captured on winter nights, where the lack of sufficient lighting renders cameras ineffective. The two sequences are collected onboard a mobile platform equipped with multi-modal sensor measurements, including color images from the Intel RealSense L515 LiDAR camera, point clouds from the solid-state LiDAR Livox Avia, and synchronized IMU data from the built-in IMU (model BMI088) of Livox Avia operating at 200 Hz.

\begin{figure}[t]
    \centering
    \includegraphics[width=1\linewidth]{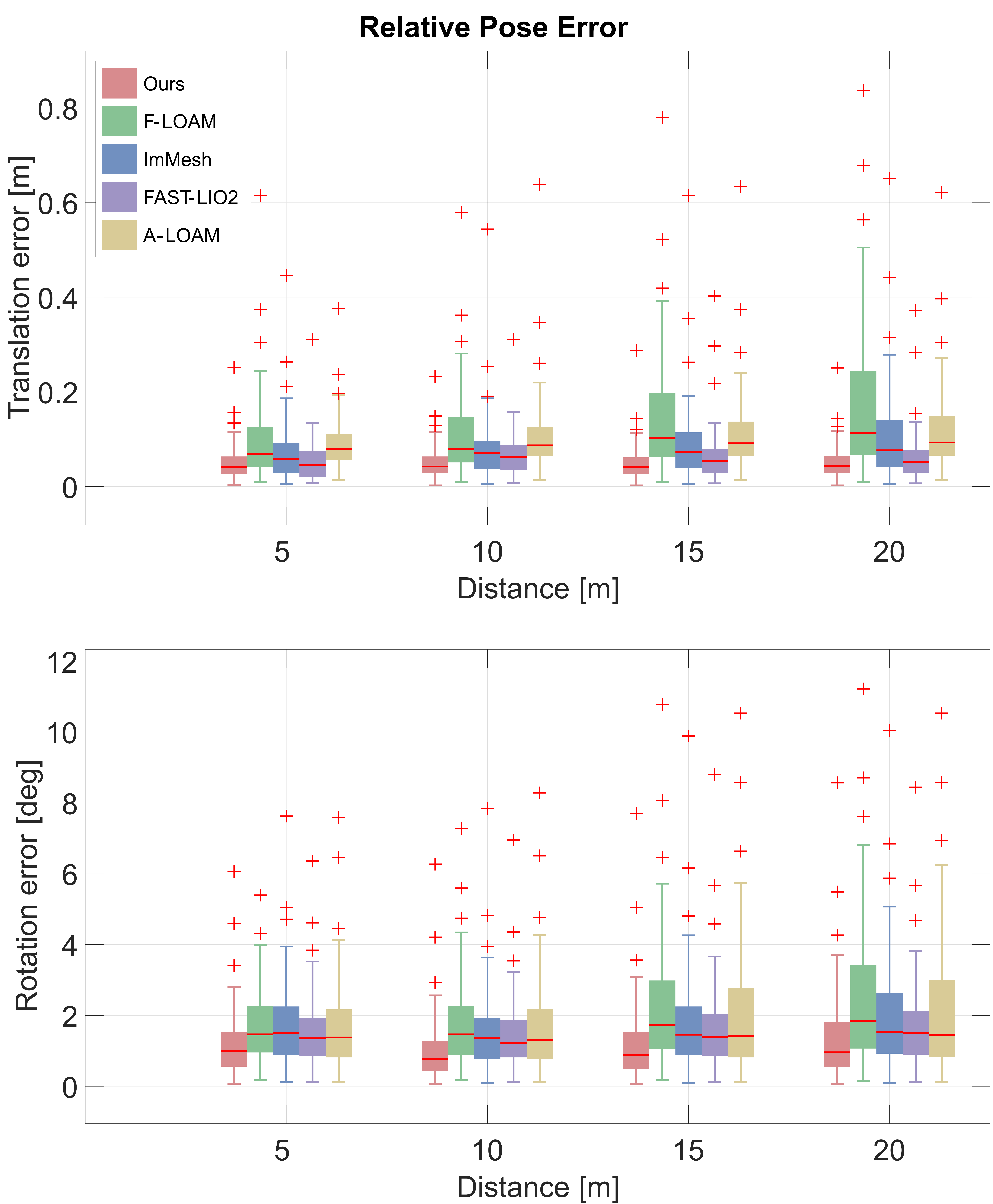}
    \caption{Relative pose error \cite{6248074} in Forest01. Two plots are relative errors in translation and rotation, respectively.}
    \label{fig:7}
\end{figure}
\begin{figure}[t]
    \centering
    \includegraphics[width=1\linewidth]{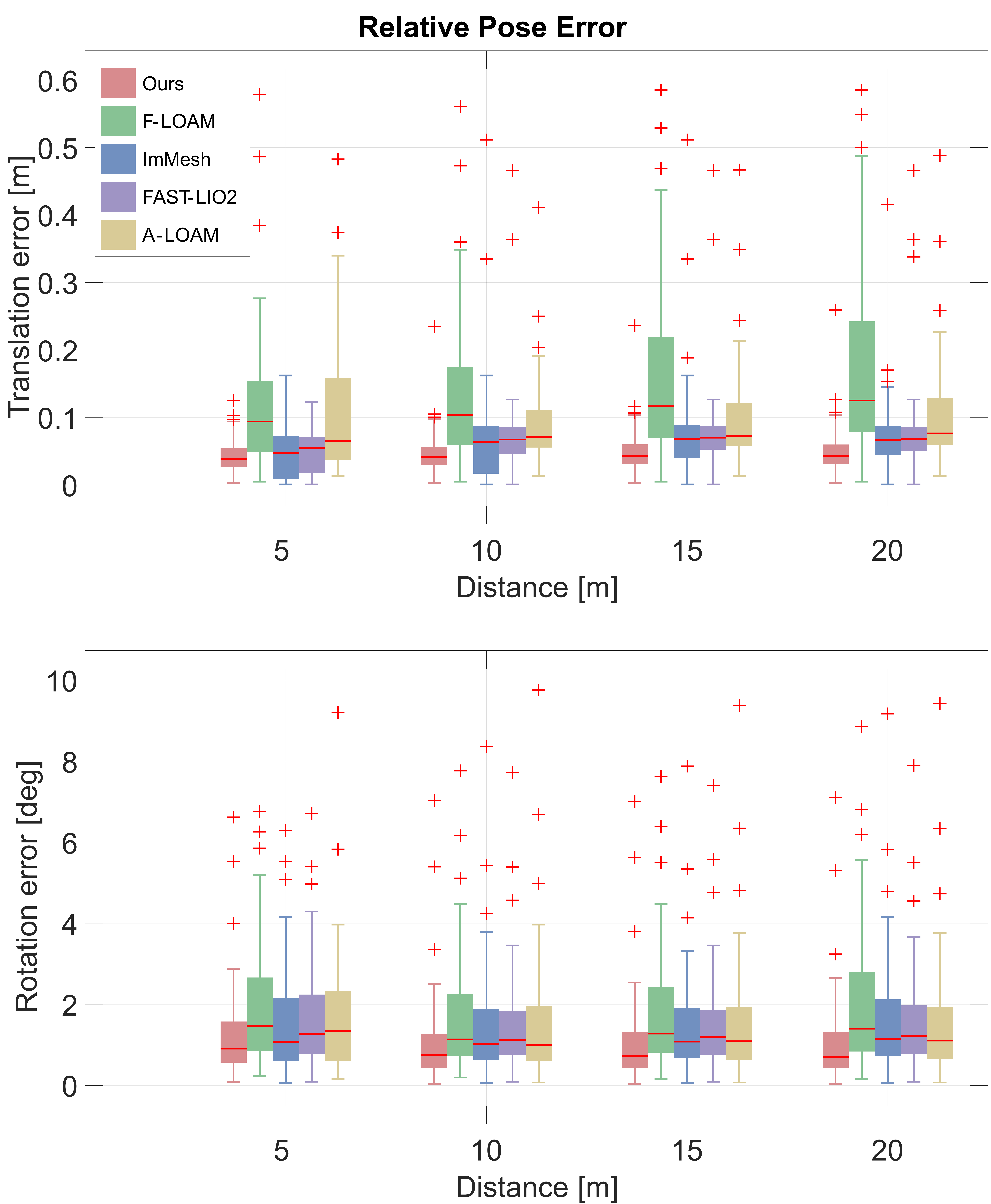}
    \caption{Relative pose error \cite{6248074} in Forest02. Two plots are relative errors in translation and rotation, respectively.}
    \label{fig:8}
\end{figure}
\begin{table}
    \centering
    \caption{RMSE of ATE \cite{6385773} in Meters}
    \label{tab:1}
    \begin{tabular}{lccc}
        \toprule
        \textbf{Methods} & \textbf{Forest01} & \textbf{Forest02} & \textbf{Campus01}\\
        \midrule
        \textbf{Ours} &  \textbf{0.063} &  \textbf{0.050} & \textbf{0.636}\\
        \textbf{F-LOAM \cite{9636655}} & 0.382 & 0.465 & 0.692\\
        \textbf{IMMESH \cite{10304337}} & 0.435 & 0.060 & 0.809\\
        \textbf{FAST-LIO2 \cite{9697912}} & \textbf{0.063} & 0.053 & 0.768\\
        \textbf{A-LOAM \cite{zhang2017low}} & 0.679 & 0.624 & 2.302\\
        \bottomrule
    \end{tabular}
\end{table}
\begin{figure}[t]
    \centering
    \includegraphics[width=\linewidth]{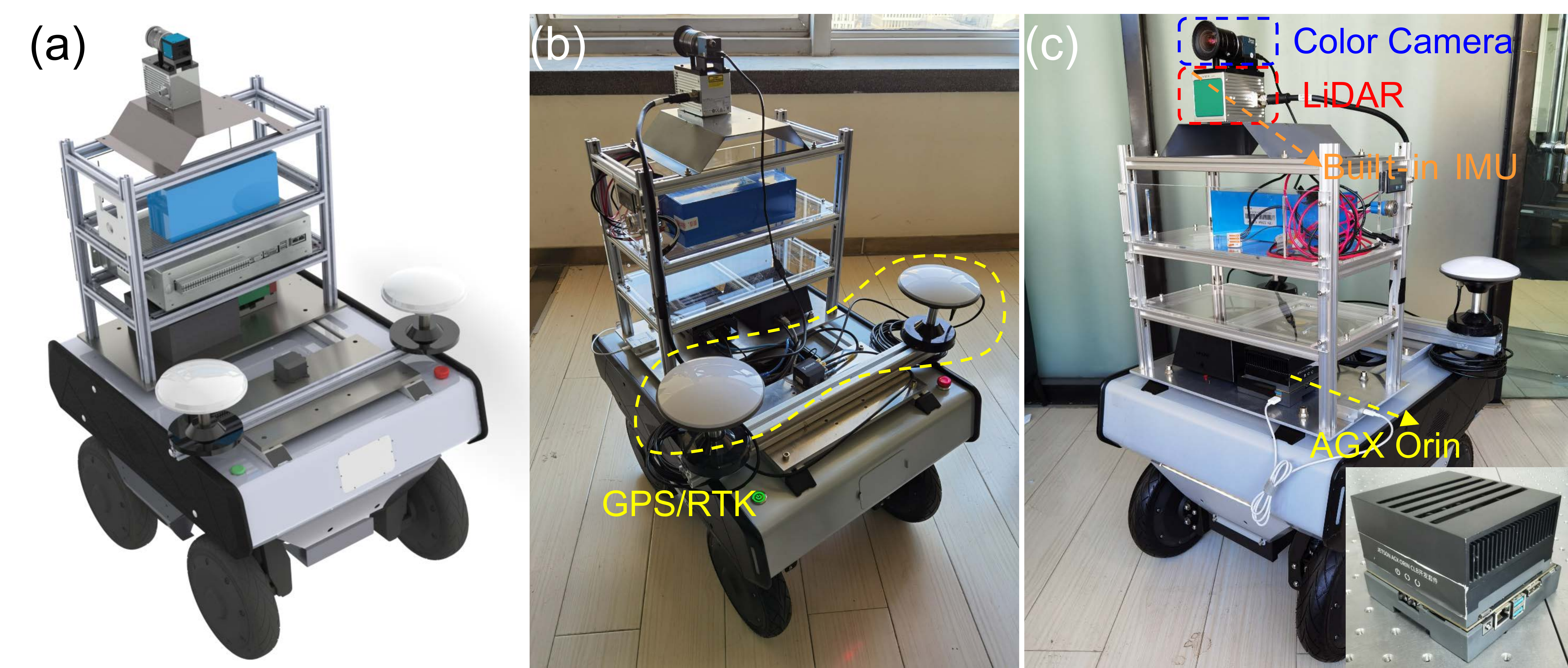}
    \caption{(a) A 3D rendered model of our whole device. (b) A rear view of the device, which integrates a GPS/RTK system for ground-truth trajectory measurement. (c) A front view of the device, where our minimal setup consists of a Livox Avia (with IMU), a camera and a Jetson AGX Orin.}
    \label{fig:9}
\end{figure}
In this experiment, we compare our method with F-LOAM \cite{9636655}, ImMesh \cite{10304337}, FAST-LIO2 \cite{9697912} and A-LOAM \cite{zhang2017low}. For the sequence Forest01, the trajectories and the map produced by our method are illustrated in Fig. 5. Disturbed by correspondence outliers from unstable features such as tree leaves, the trajectory of F-LOAM deviates significantly from ground truth throughout the entire path. A-LOAM and ImMesh begin to exhibit substantial drifts and distortion starting from the first and second turns, respectively. The feature extraction module of A-LOAM is developed only for spinning LiDARs, which is the main reason for its poor performance in both this experiment and Experiment-2. For consistency, IMU data is only used in FAST-LIO2. Our method provides very accurate trajectory results comparable to FAST-LIO2, even without using any IMU data. These observations are further supported by Table~\ref{tab:1}, which summarizes the root-mean-squared errors (RMSE) in meters across multiple sequences for each algorithm, evaluated using the absolute trajectory errors (ATE) \cite{6385773}. The relative pose errors (RPE) evaluated by \cite{6248074} are presented in Fig.~\ref{fig:7}. In the error plot, our method consistently outperforms the other methods, with lower translation and rotation errors. Similar results are observed for the sequence Forest02, as depicted in Fig.~\ref{fig:6} and Fig.~\ref{fig:8}. Given the absence of effective color information at night, the results above demonstrate that our robust error metric based on the Welsch's function is highly resilient to a large number of positional correspondence outliers in forest environments.
\subsection{Experiment-2: High accuracy odometry in a large-scale outdoor campus environment}
\emph{1) Real-world Experiment:} In this experiment, we aim to develop an automated guided vehicle (AGV) to replace labor-intensive manufacturing. The AGV is designed to carry out daily tasks such as routine inspections. As shown in Fig.~\ref{fig:9}, the minimal system of CAR-LOAM includes a Livox Avia LiDAR, a MER2-230-168U3C color camera with a matching lens, an NVIDIA Jetson AGX Orin Developer Kit as the computation platform, and power supplies. Our minimal system is mounted on the Yuhesen FW-01 mobile platform with power supplies, and the YESENSE YIN660-D GPS/RTK system provides ground-truth measurements, ensuring centimeter-level positioning accuracy when the signal is strong. The Field of View (FoV) of the LiDAR is $70.4^\circ \times 77.2^\circ$ while the FoV of the camera is $70.0^\circ \times 47.6^\circ$. The built-in IMU of Avia operates at 200 Hz. During the test, the mobile platform travels at an average speed of 1.4 m/s over a total distance of 200 meters within the campus of Beijing Institute of Technology. The mapping results are shown in Fig.~\ref{fig:1}.
\begin{figure}[t]
    \vspace{0.3cm}
    \centering
    \includegraphics[width=1\linewidth]{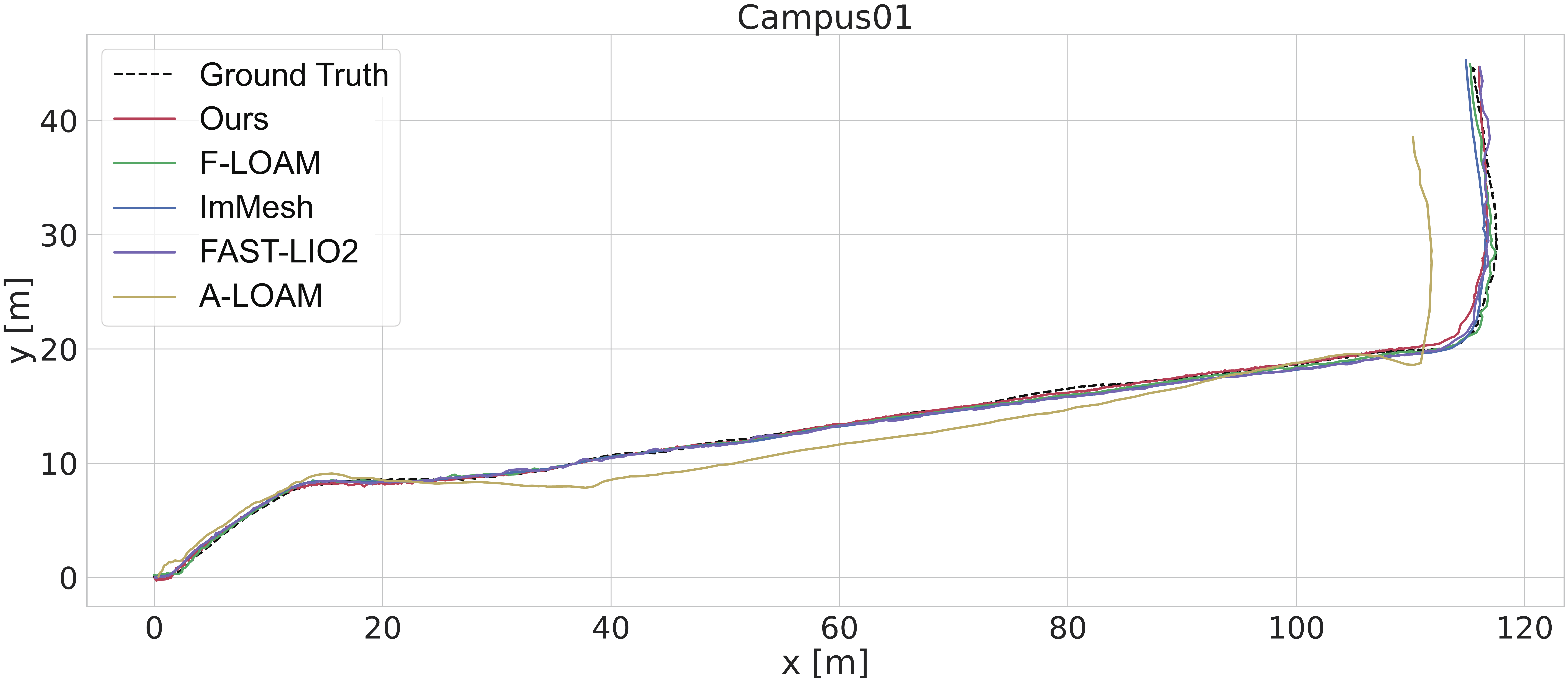}
    \caption{Trajectories in Campus01.}
    \label{fig:10}
\end{figure}

\begin{figure}[t]
    \centering
    \includegraphics[width=1\linewidth]{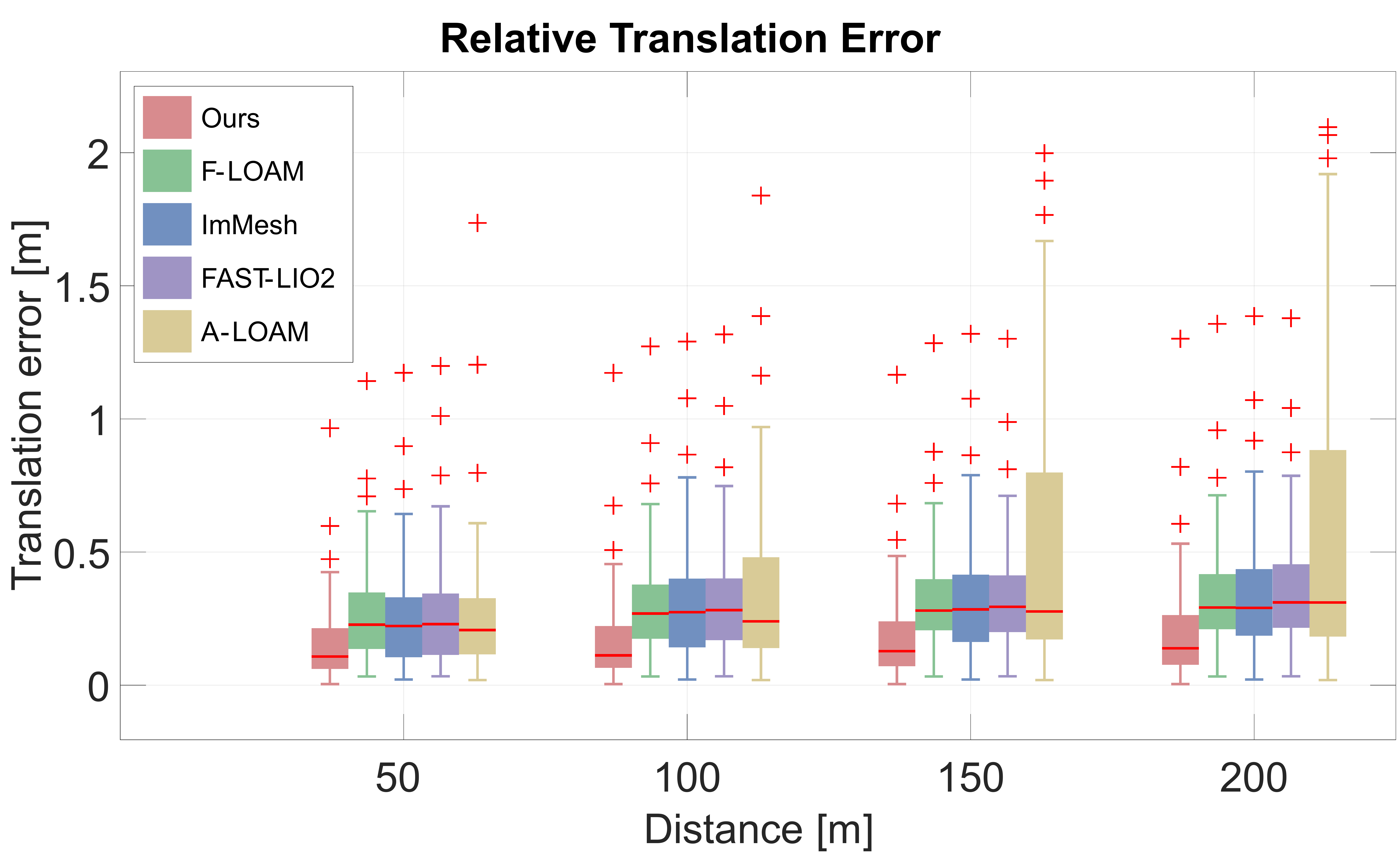}
    \caption{Relative translation error \cite{6385773} in Campus01.}
    \label{fig:11}
\end{figure}

To evaluate the performance of our method, we compare it with the four previously mentioned algorithms on the Campus01 dataset we collected. The absolute trajectory errors of the five algorithms are also shown in Table.~\ref{tab:1}. The trajectories are shown in Fig.~\ref{fig:10}. We then evaluate their relative translation errors (RTE) \cite{6385773}, and the error plot is presented in Fig.~\ref{fig:11}. Due to the suppression of color and positional correspondence outliers, our method maintains higher accuracy during long-distance operation.

\emph{2) Ablation study:} To further demonstrate the performance of the proposed perceptually uniform color difference weighting strategy and the robust error metric via the Welsch's function, we compare the results of different approaches on our Campus01 dataset, as shown in Table~\ref{tab:2}. It can be seen that the two methods reduce erroneous scan matching to improve accuracy by addressing color correspondence outliers and positional correspondence outliers, respectively. Moreover, in general scenarios, combining both methods yields better performance, with ATE reduced by 34.9\% and RTE reduced by 35.5\%.
\begin{table}[ht]
    \vspace{0.3cm}
    \centering
    \caption{RMSE of ATE and RTE \cite{6385773} in meters in the ablation study.}
    \label{tab:2}
    \begin{tabular}{cccc}
        \toprule
        \textbf{Color weight} & \textbf{Welsch's} & \textbf{ATE} & \textbf{RTE} \\
        \midrule
        w/ & w/ & \textbf{0.636} & \textbf{0.331}\\
        w/ & w/o & 0.718 & 0.351\\
        w/o & w/ & 0.747 & 0.332\\
        w/o & w/o & 0.977 & 0.513\\
        \bottomrule
    \end{tabular}
\end{table}
\subsection{Experiment-3: High precision 3D reconstruction of a large building}
\begin{figure}[t]
    \centering
    \includegraphics[width=1\linewidth]{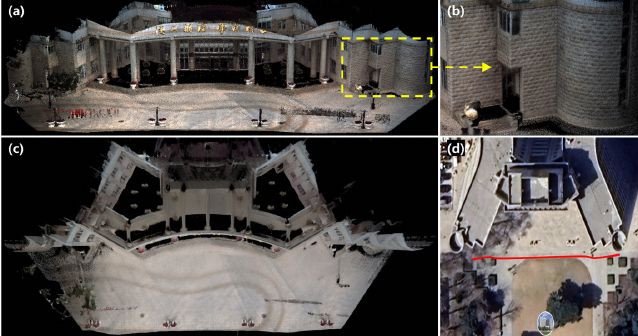}
    \caption{
        Colored 3D reconstructed point cloud model of the central teaching building.
        (a) Front view of the model.
        (b) A close-up of (a), showing the density, rich textures, and high fidelity of the reconstruction.
        (c) Top view of the model, demonstrating the clear boundaries achieved in our reconstruction.
        (d) Corresponding satellite map, with the red path indicating the trajectory of the AGV during reconstruction.
    }
    \label{fig:12}
\end{figure}
\begin{figure}[!t]
   \centering
   \includegraphics[width=\linewidth]{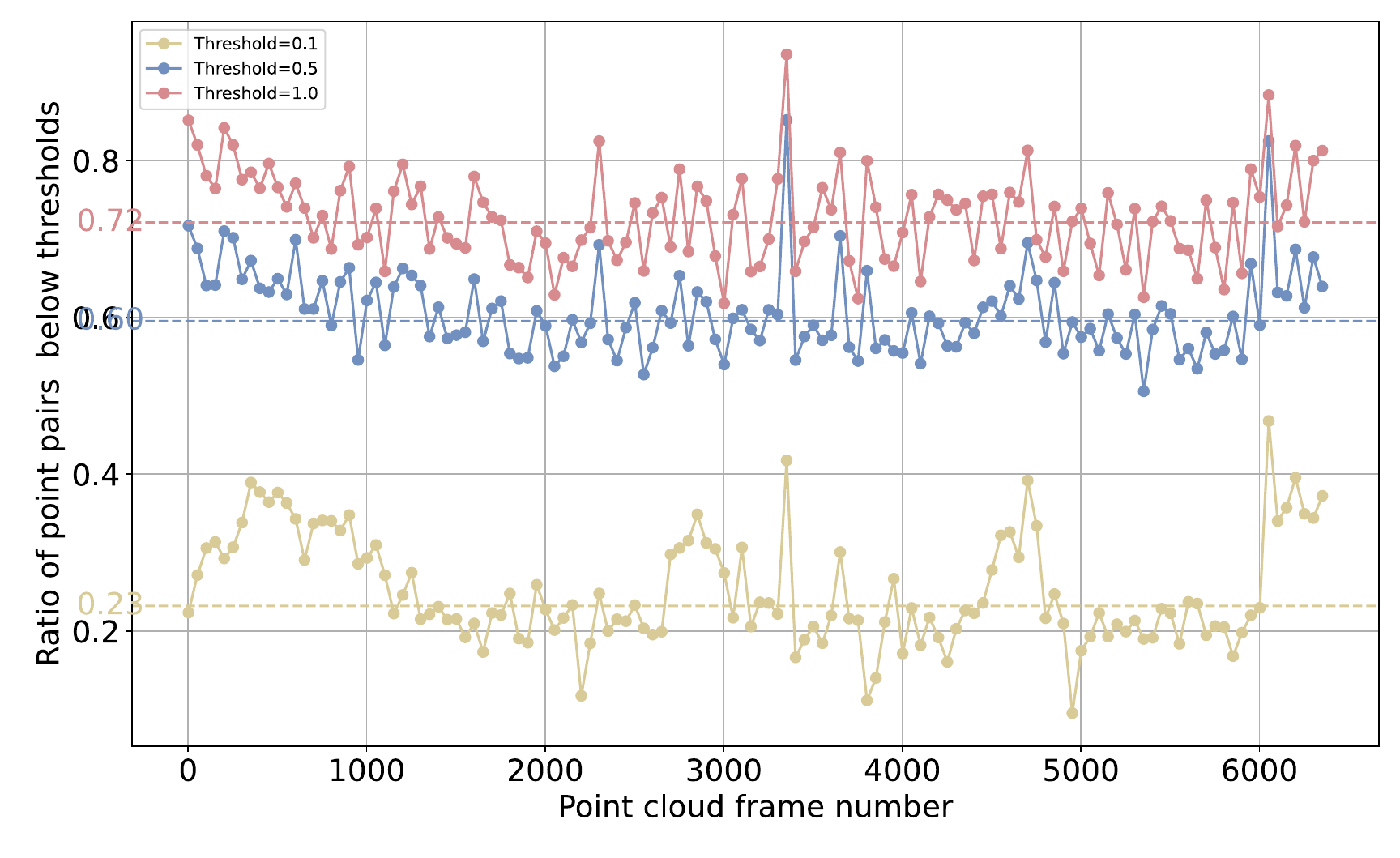}
   \caption{Ratio of point pairs with distances below thresholds in Experiment 3. Dashed lines represent the average percentage across all frames for each threshold.}
   \label{fig:13}
\end{figure}
In this experiment, we aim to apply our method to the field of 3D reconstruction. We continue to use the equipment in Experiment-2 to perform a 3D reconstruction of the eastern side of the central building on campus. During the movement, the LiDAR and camera are consistently oriented towards the central teaching building. Through the lateral movement capability of the Yuhesen FW-01, the vehicle maintains a trajectory that is almost perpendicular to the scanning direction. The mapping results are shown in Fig.~\ref{fig:12}, where the generated colored 3D model exhibits dense point clouds, rich textures, clear boundaries and high fidelity.

To evaluate the reconstruction results indirectly, each of the 6,363 registered point cloud frames is paired with its previous frame (if available) to form a frame pair. For each frame pair, we calculated the percentage of nearest neighbor point pairs with distances below thresholds of 0.1 mm, 0.5 mm and 1 mm. Then, the results are plotted as a curve graph shown in Fig.~\ref{fig:13}. For clarity, one point is plotted on the graph for every 40 frames.
The graph shows that, on average, across all frame pairs, 23\% of nearest neighbor point pairs have distances below 0.1 mm, 60\% below 0.5 mm, and 72\% below 1 mm. This indicates high inter-frame consistency of the point clouds, leading to highly accurate reconstruction.
\section{CONCLUSION}
The key of our framework is the use of color weighting and a robust error metric to exclude color and positional outliers, thereby reducing erroneous registrations in pose estimation. Dataset experiments demonstrate a significant reduction in drift in forests filled with unstable features. Real-world experiments show that, in general environments, our method outperforms current state-of-the-art approaches. The ablation study indicates that using our strategy improves localization accuracy by 34.9\% (ATE) and 35.5\% (RTE). 
Although our method is applicable to most general scenarios, it may perform poorly in featureless degenerated environments or those with similar or lack of color. In future work, we envision integrating IMU with our system to further enhance its accuracy and robustness.




\bibliographystyle{IEEEtran}
\bibliography{references}

\end{document}